\documentclass{article} % For LaTeX2e
\usepackage[top=1in, left=1in, right=1in, bottom=1in]{geometry}
\usepackage[square,sort,comma,numbers]{natbib}
\usepackage{amsmath,amsfonts,bm}

\def\eqref#1{equation~\ref{#1}}
\def\Eqref#1{Equation~\ref{#1}}
\def\1{\bm{1}}

\def\mU{{\bm{U}}}
\def\mV{{\bm{V}}}
\def\mW{{\bm{W}}}
\def\mX{{\bm{X}}}

\DeclareMathAlphabet{\mathsfit}{\encodingdefault}{\sfdefault}{m}{sl}
\SetMathAlphabet{\mathsfit}{bold}{\encodingdefault}{\sfdefault}{bx}{n}

\usepackage{graphicx}
\usepackage{hyperref}
\usepackage{subcaption}
\usepackage{url}
\usepackage{pifont}
\usepackage{authblk}
\usepackage{multirow}
\usepackage{tabularray}
\usepackage{wrapfig,lipsum,booktabs}
\newcommand{\wta}[1]{\texttt{FFSplit\!\!} #1}
\newcommand{\printfnsymbol}[1]{%
  \textsuperscript{\@fnsymbol{#1}}%
}

\title{FFSplit: Split Feed-Forward Network For Optimizing Accuracy-Efficiency Trade-off in Language Model Inference}

\author[1]{Zirui Liu\thanks{Work done during the internship at LinkedIn, zl105@rice.edu}}
\author[2]{Qingquan Song}
\author[2]{Qiang Charles Xiao}
\author[2]{Sathiya Keerthi Selvaraj}
\author[2,3]{Rahul Mazumder}
\author[2]{Aman Gupta}
\author[1]{Xia Hu}
\affil[1]{Rice University}
\affil[2]{LinkedIn Corporation}
\affil[3]{Massachusetts Institute of Technology}
\begin{document}

\maketitle

\begin{abstract}

The large number of parameters in Pretrained Language Models enhance their performance, but also make them resource-intensive, making it challenging to deploy them on commodity hardware like a single GPU.
Due to the memory and power limitations of these devices, model compression techniques are often used to decrease both the model's size and its inference latency. This usually results in a trade-off between model accuracy and efficiency. 
Therefore, optimizing this balance is essential for effectively deploying LLMs on commodity hardware.
A significant portion of the efficiency challenge is the Feed-forward network (FFN) component, which accounts for roughly $\frac{2}{3}$ total parameters and inference latency.
In this paper, we first observe that only a few neurons of FFN module have large output norm for any input tokens, a.k.a. heavy hitters, while
the others are sparsely triggered by different tokens.
Based on this observation, we explicitly split the FFN into two parts according to the heavy hitters. We
improve the efficiency-accuracy trade-off of existing compression methods by allocating more resource to FFN parts with heavy hitters.
In practice, our method can reduce model size by 43.1\% and bring $1.25\sim1.56\times$ wall clock time speedup on different hardware with negligible accuracy drop.
\end{abstract}

\section{Introduction}

Pre-trained language models (LMs) with transformer architecture have achieved remarkable success in numerous natural language processing (NLP) tasks \citep{transformer, devlin2018bert, t5, gpt3}.
Recent research has clearly shown that increasing the number of parameters in pre-trained language models significantly enhances their performance~\citep{kaplan2020scaling}.
However, these models, equipped with billion-scale parameters, come with high costs in terms of storage, memory, and inference latency. 
% There exists a significant disparity exists between the memory requirements of pre-trained LMs and the capacity of current hardware, particularly GPUs.

This has motivated a growing interest in model compression techniques, aiming to make the models more compact and efficient for real-world applications \citep{xu2023compress, cofi, wtacrs, frantar2022gptq}.
These model compression methods can be roughly divided into three categories.
First, some works have suggested pruning the large pre-trained models to identify a more efficient and accurate subnetwork \citep{frantar2023sparsegpt, cofi}.
Second, another research line quantizes the model weights into lower numerical precision \citep{frantar2022gptq, xiao2023smoothquant, lin2023awq}.
Third, some other works try to apply low-rank decomposition to the weight matrix \citep{chen2021drone, zhao2023inrank}.
All these three methods essentially trade off model quality to reduce the time and/or memory complexity.
This results in a trade-off between accuracy and efficiency.

Each transformer layer consists of a multi-head self-attention (MHA) part, and
a feed-forward network (FFN) part \citep{transformer}.
We note that FFN is the key efficiency bottleneck because it takes $\frac{2}{3}$ total parameters and inference latency \citep{dejavu}.
In parallel, prior studies have observed a "heavy hitter" phenomenon in ReLU-based language models' FFN modules \citep{lazyneuron, dejavu}. 
This means only a few neurons\footnote{To avoid create fusion, ``neuron'' in this paper is equivalent to the output dimension of the first FFN layer.} of FFNs are have non-zero outputs after ReLU for almost all tokens, while the rest neurons are sparsely activated.
This observation indicates that we waste many computation resource on non-important neurons. 
However, we note that in practice, the dominant language models are based on GeLU or its variants \citep{devlin2018bert, swiglu}, which inherently don't showcase such activation sparsity. 
As a result, this "heavy hitter" phenomenon remains largely unexplored for mainstreaming language models.
In view of such, we ask: \textit{Whether ``heavy hitters'' exist in non-ReLU based transformers? If so, can we leverage this observation to improve the accuracy-efficiency trade-off of compressed FFN module?}

This paper makes an attempt in providing a positive answer to the above questions.
Specifically, we first found that for non-ReLU based transformers, ``heavy hitters'' still exist and matter for the model performance.
Namely, we found that only a few neurons of FFN module have large output norm for any input tokens, while the others are sparsely triggered by different tokens.
Based on this, we propose to identify the set of ``heavy hitter'' neurons by going through a small set of training samples. 
Then as shown in Figure \ref{fig: diag}, we explicitly split the FFN into two parts according to the heavy hitters.
We allow more resource to the FFN part with heavy hitters when applying model compression methods.
In this way, we improve the efficiency-accuracy trade-off of existing compression methods.
In summary, our contributions are:

\begin{itemize}
    \item We found that only a few neurons of FFN module have large output norm for any input tokens, while the others are sparsely triggered by different tokens.
    \item Based on the observation, we explicitly split the FFN into two parts according to the heavy hitters.
    We improve the efficiency-accuracy trade-off of existing compression methods by allocating more resource to FFN parts with heavy hitters.
    \item In practice, our method can reduce model size by 43.1\% and bring $1.25\sim1.56\times$ wall clock time speedup on different hardware with negligible accuracy drop.

\end{itemize}
\section{Background and Motivation}

A Transformer network \citep{transformer} is
composed of several layers and each layer consists
of a multi-head self-attention (MHA) part, and
a feed-forward network (FFN) part.
In this paper, we use the following notations for clarity:
% $b$ represents the batch size.
% $s$ indicates the length of the input sequence.
$d$ is the hidden dimension.
$d_{ff}$ refers to the hidden dimension of the FFN layer.
$l$ denotes the total number of transformer layers.
Typically, we have $d_{ff} = 4d$  \citep{transformer}.
Within the $i^{\text{th}}$ transformer layer,  we use $\mW_Q^i, \mW_K^i, \mW_V^i, \mW_O^i\in\mathbb{R}^{d\times d}$ to represent the weight matrices of the Query, Key, Value, and Output layers of the MHA, respectively.
$\mU^i \in\mathbb{R}^{d\times d_{ff}}$ and $\mV^i \in\mathbb{R}^{d_{ff}\times d}$ are the up-projection and down-projection layer of the FFN, respectively.
Usually speaking, \textbf{the FFN part takes $\frac{2}{3}$ total parameters} \citep{dejavu} (embedding is excluded). The FFNs can be expressed as
\begin{equation}
    \text{FFN}(\mX) = \sigma(\mX\mU)\mV, \nonumber
\end{equation}
where $\mX\in\mathbb{R}^{s\times d}$ is the input tensor and $s$ is the sequential length. $\sigma$ is the activation function, e.g., GeLU \citep{hendrycks2016gaussian}. Following tiled matrix mulplication, we can decompose the FFNs as follows:
\begin{equation}
\label{eq:ffn_decompose}
    \text{FFN}(\mX)=\sum_{j=1}^{d_{ff}}\sigma(\mX\mU_{:,j})\mV_{j,:},
\end{equation}
where $\mU_{:,j}$ is the $j^{\text{th}}$ column of $\mU$ and $\mV_{j,:}$ is the  $j^{\text{th}}$ row of $\mV$, respectively.
\Eqref{eq:ffn_decompose} means $\text{FFN}(\mX)$ can be expressed as the sum of $d_{ff}$ rank-one matrix, where each rank-one matrix is the outer production between one column of $\sigma(\mX\mU)$ and one row of $\mV$.

Previous studies have shown that in ReLU-based language models, e.g., OPT \cite{opt} and T5 \citep{t5}, a subset of the $d_{ff}$ neurons are ``heavy hitters'' \citep{lazyneuron, dejavu}.
Specifically, a few neurons have non-zero outputs after ReLU for almost all tokens, while the rest neurons are sparsely activated. 
Yet, we note that in practice, the dominant language models are based on GeLU \cite{devlin2018bert} or SwiGLU \cite{swiglu}.
The key difference between the GeLU family and the ReLU activation function lies in the ability of GeLU (and its variants) to give non-zero outputs for small negative values.
Given this characteristic, \textbf{we hypothesize that ``heavy hitter'' neurons also exist in non-ReLU based language models.} However, "heavy hitter" should be defined based on the norm, considering GeLU's potential non-zero output for small negative inputs.
Mathematically, this can be understood as there being some $j\in [d_{ff}]$ for which $\|\sigma(\mX\mU_{:,j})\|_F$ is large for any input tensor $\mX$, while the norms of the rest neurons remain small.
% Specifically, a few neurons have large norms for nearly all tokens, while the rest neurons are sparsely activated. Mathematically, this can be understood as there being some $j\in [d_{ff}]$ for which $\|\sigma(\mX\mU_{:,j})\|_F$ is large for any input tensor $\mX$, while the norms of the rest neurons remain small.
If we can identify the set of ``heavy hitters'' neurons, denoted as $\texttt{h}_2$, then we can explicitly decouple the original FFNs into two separate parts:

\begin{align}
    \text{FFN}(\mX) &= \sum_{j\in \texttt{h}_2}\sigma(\mX\mU_{:,j})\mV_{j,:} + \sum_{j\notin \texttt{h}_2}\sigma(\mX\mU_{:,j})\mV_{j,:} \nonumber\\
    &= \text{FFN}_1(\mX) + \text{FFN}_2(\mX), \label{eq: ffn decompose}
\end{align}

where $\text{FFN}_1$ is the sub-FFN specified by the heavy hitters, while  $\text{FFN}_2$ is the sub-FFN with the remain neurons. 
\textbf{Our goal is to reduce the model's size while achieving faster inference in terms of wall-clock time}.
Below we discuss advantages of explicitly splitting FFNs to achieve this goal:

\textbf{Why splitting FFNs into two parts:} The main motivation of splitting FFNs into two separate parts is two-fold: 
\textbf{(1)} It still use dense matrix format to do the computation, and thus to be hardware-friendly for potentially obtaining wall-clock time speedup; 
\textbf{(2)} Any compression technique can be applied over this formulation. We provide a finer granularity when balancing the trade-off between efficiency and accuracy. Specifically, we hypothesize that few "heavy hitters" play a crucial role in determining model performance.
If our hypothesis is true, then $\text{FFN}_1$ emerges as a compact yet powerful component. 
During model compression, we should allocate more resources to $\text{FFN}_1$ than $\text{FFN}_2$. In the next Section, we will validate our hypothesis.

\section{Related Work}

In this section, we will begin by introducing the efficiency bottleneck of LM inference. 
Then we will introduce current approximation approaches that are designed to reduce the computation and memory overhead and improve LLM inference latency. 
% Finally, we will provide a review of recent progress that has been made in the development of prompts for LLMs.

\subsection{Efficiency Bottleneck of LM Inference}
LLMs use a decoder-only, autoregressive method where tokens are generated sequentially, with each token depending on prior results. For example, models like GPT, as cited in \cite{radford2018improving, radford2019language, brown2020language}, operate on this principle. Recent research by \cite{liu2023DejaVu} on the OPT-175B models' inference process reveals that: (1) token generation is the primary cause of inference latency, and (2) during token generation, the Multilayer Perceptron (MLP) has higher I/O and computation delays compared to attention blocks. Although system-level optimizations, as mentioned in \cite{sheng2023high,mlcllm, web-llm, vllm}, can speed up LLM inference times, they don't directly address the computational and memory I/O challenges in the LLM inference process.

\subsection{Approximation in LM Inference}
Beyond system-level optimizations, there are two main strategies to decrease both computational and memory I/O requirements, thus reducing inference latency. (1) Sparse Modeling: This involves selecting specific weights in certain layers to lessen both computational and memory I/O demands, as seen in \citep{frantar2023sparsegpt,liu2023DejaVu,liu2023winner}. These methods are akin to pruning techniques described in \citep{he2018amc,kwon2022fast,hubara2021accelerated}. Given LLMs' vast number of parameters, sparsification is usually applied layer by layer. Yet, the resulting sparse LLM can differ notably in its final inference predictions, often leading to reduced accuracy compared to the original LLM. (2) Quantization: This entails compressing the trained weight values of LLMs into fewer bits, as detailed in \citep{nagel2020up,dettmers2022llm,xiao2022smoothquant,frantar2022gptq,xu2023compress}. Studies indicate that int8 quantization can closely approximate the original LLMs' predictive capabilities \citep{dettmers2022llm}. However, further reducing the bit count can lead to a substantial accuracy drop.

% \subsection{Prompt for LLMs} 
% LLMs are known for their in-context learning ability, allowing them to generalize to unseen tasks without additional fine-tuning \citep{brown2020language}.
% Specifically, LLMs are controlled through user-provided natural language specifications of the task, or \emph{prompts}, which illustrate how to complete a task. 
% In this paradigm, we do not enforce modifications on the LLMs themselves. Instead, we focus on adapting the inputs to the LLMs for better predictive performance in downstream tasks. A typical strategy is to insert tokens before the input sequence to affect the attention mechanism. It has been shown in \citep{brown2020language} that prompt engineering enables LLMs to match the performance of fine-tuned language models on a variety of language understanding tasks. Moreover, \citep{lester2021power} empirically indicate that there is an equivalence between modifying the input and fine-tuning the model. Furthermore, \citep{su2022transferability} studies the transferability of prompts across similar datasets or even tasks. Since then, we have witnessed the growth of prompt tuning infrastructure~\citep{ding2022openprompt}. However, we would like to emphasize that most of the current demonstrations of prompt tuning are task-specific~\citep{li2021prefix,lester2021power}. When considering efficiency, it is desirable for a prompt to exhibit transferability across various settings.
\section{Methodology}

As we previously mentioned, if ``heavy hitter'' neurons exist and are important for the model performance, then we can explicitly split FFNs into two separate parts, allocating different resources to each during compression.
In this way, we achieve a superior trade-off between the efficiency and accuracy.
In Section \ref{sec: valid}, we first validate our hyperthesis.
Then we discuss how to implement this idea in practice to obtain wall-clock time speedup in Section \ref{sec: implementation}.

\subsection{Heavy Hitter Exists and Matters for Performance}
\label{sec: valid}

\begin{figure}[h!]
    \centering
    \begin{subfigure}[h]{0.24\linewidth}
    \includegraphics[width=\linewidth]{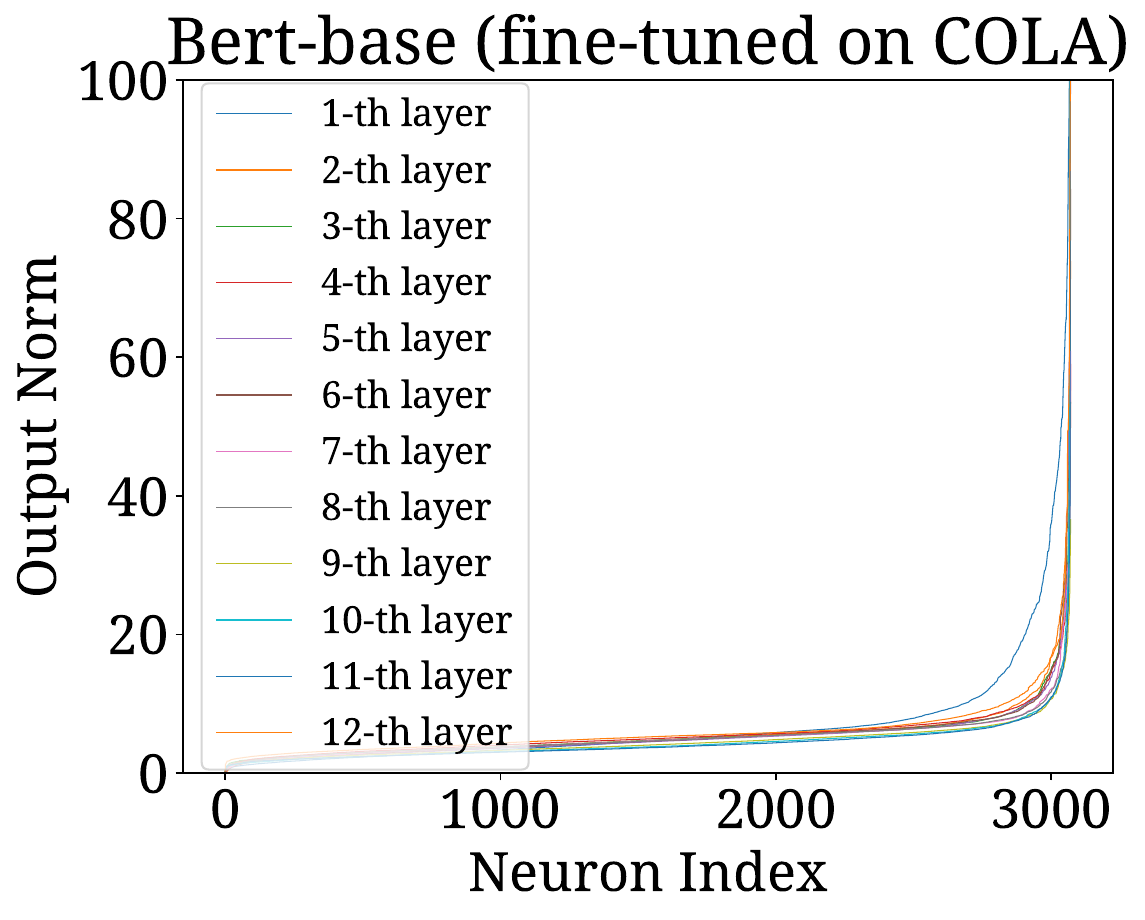}
    \end{subfigure}\hspace{-.08em}
    \begin{subfigure}[h]{0.24\linewidth}
    \includegraphics[width=\linewidth]{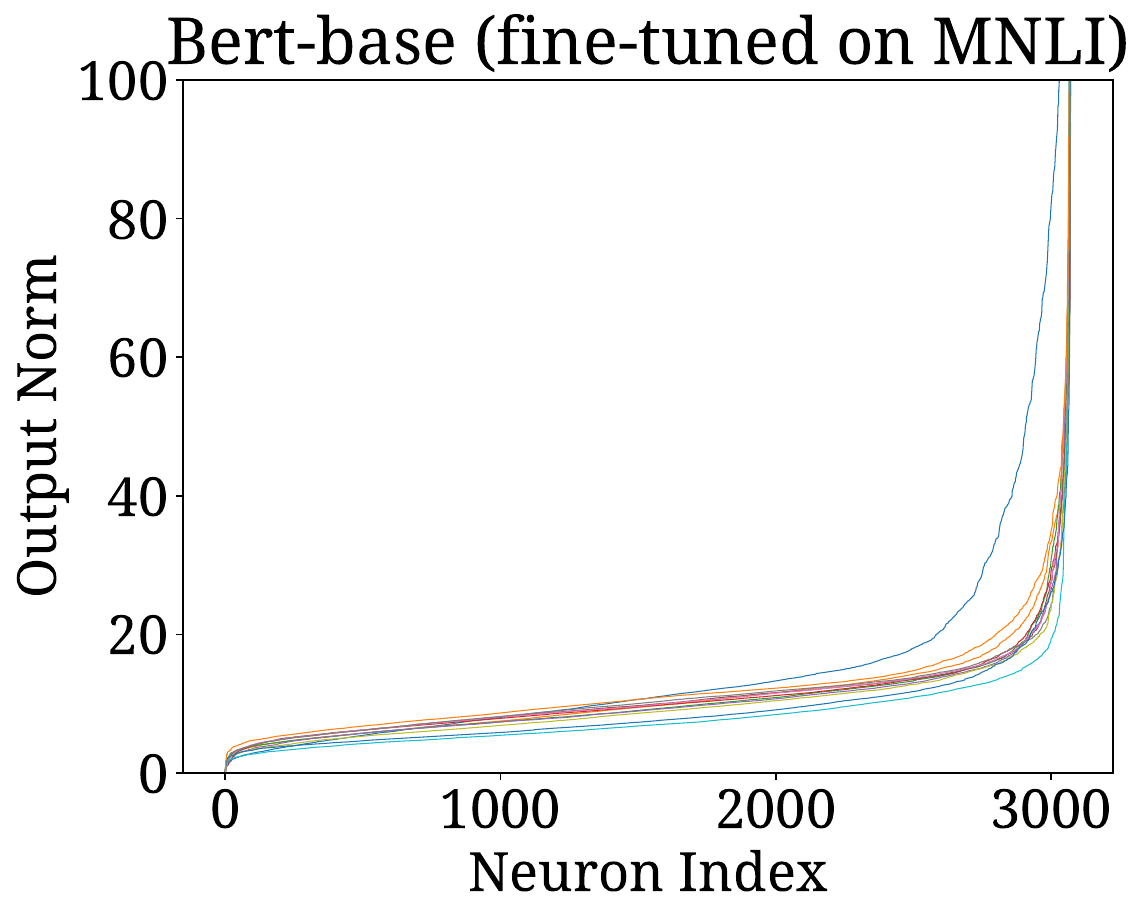}
    \end{subfigure}\hspace{-.08em}
    \begin{subfigure}[h]{0.24\linewidth}
    \includegraphics[width=\linewidth]{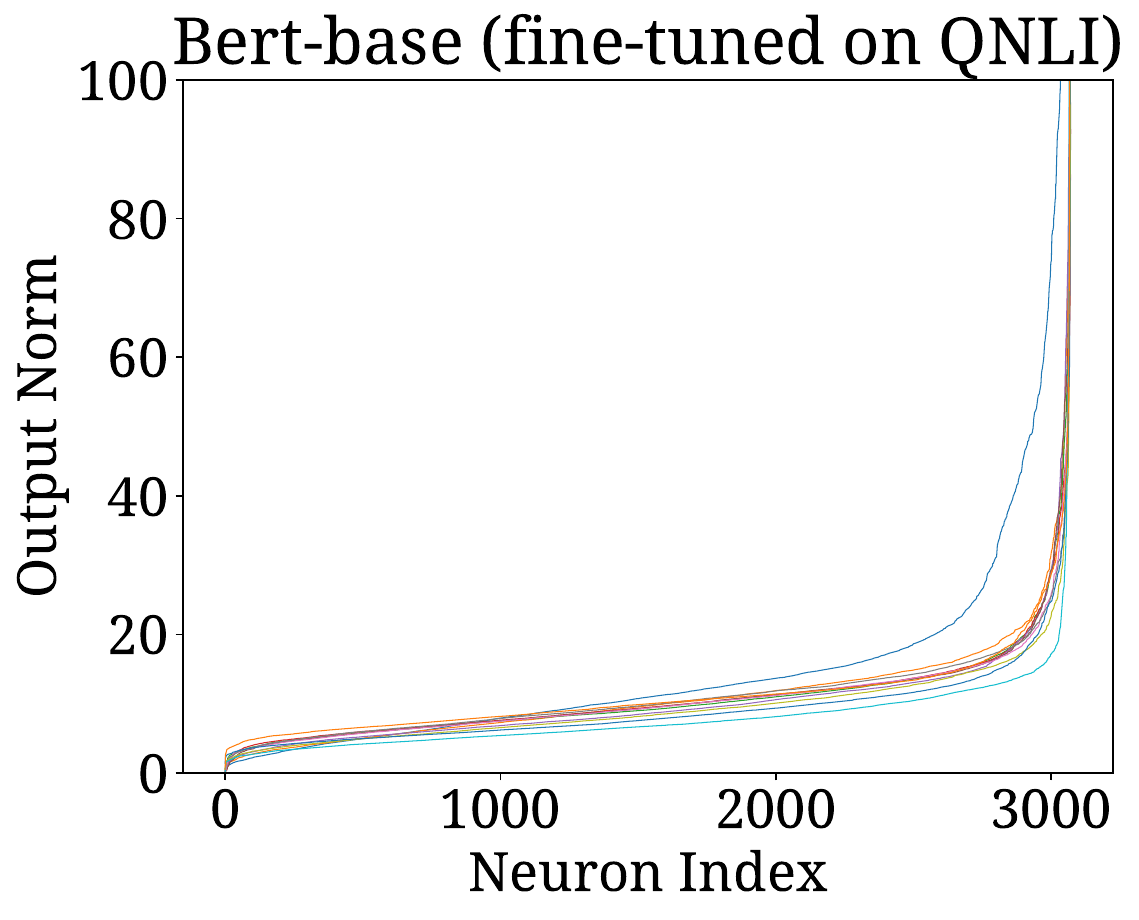}
    \end{subfigure}\hspace{-.08em}
    \begin{subfigure}[h]{0.24\linewidth}
    \includegraphics[width=\linewidth]{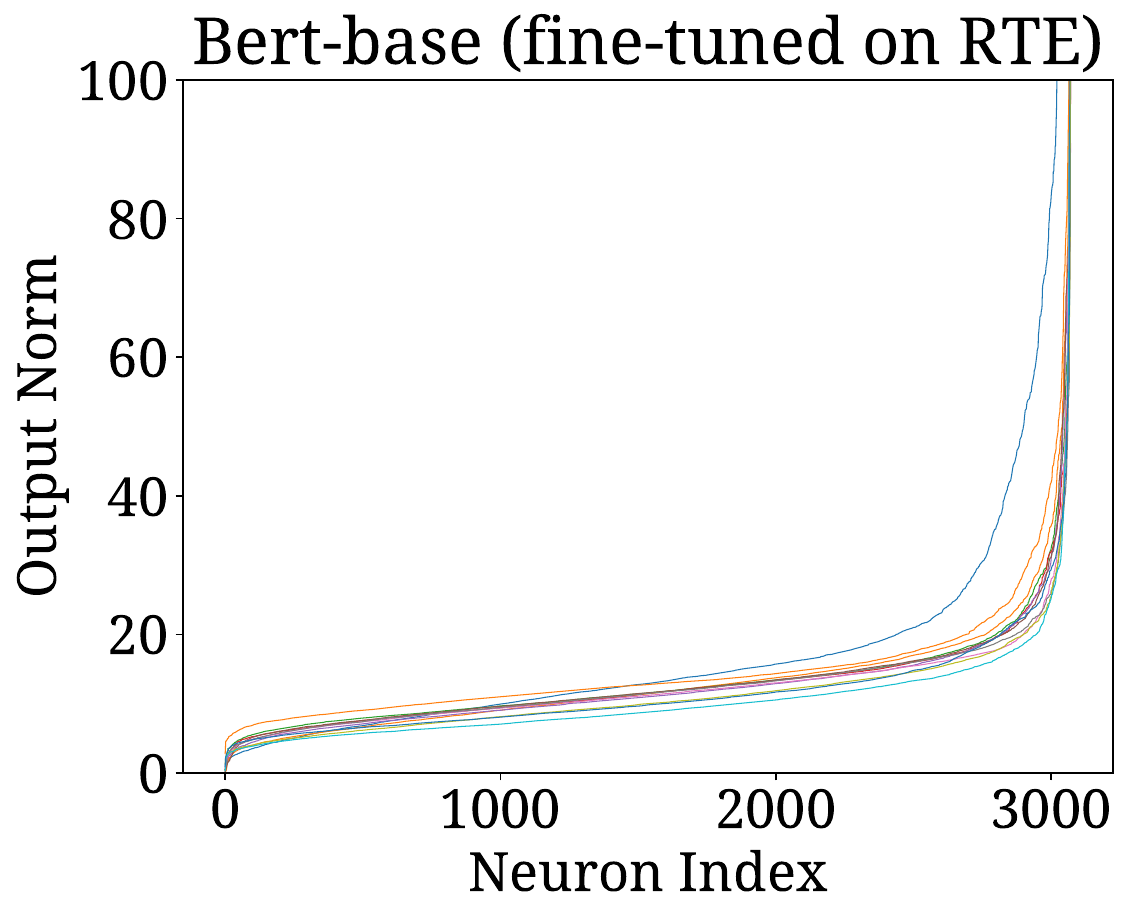}
    \end{subfigure}
    \caption{Heavy Hitter neurons also exist in GeLU-based language models.}
    \label{fig: h2 exists}
\end{figure}

In this Section, we verify the mentioned hyperthesis experimentally and mathmatically.
We first verify whether there exists heavy hitter neurons in GeLU-based language models. 
Specifically, we go through the training set with a Bert-Base \citep{devlin2018bert} model on different tasks.
Then we sort the neurons based on their output norm at each layer, as depicted in Figure \ref{fig: h2 exists}.
Neurons with the highest output norm were labeled as ``heavy hitters''.
\textit{We observe that the output norm of different neurons exhibits a long-tailed distributions, which indicates the existence of ``heavy hitters''.}

\begin{wrapfigure}{r}{0.35\textwidth}
 \vspace{-2em}
  \begin{center}
    \includegraphics[width=0.35\textwidth]{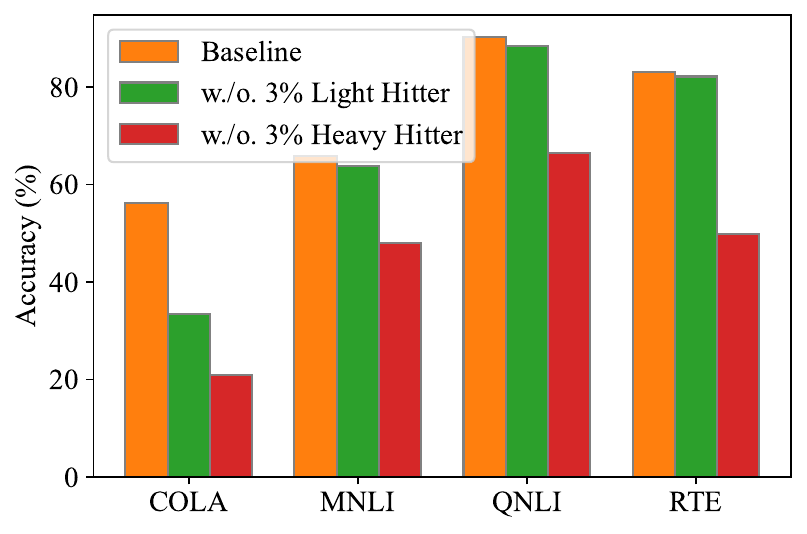}
  \end{center}
\vspace{-.5em}
    \caption{The comparison between the baseline model, the model without top 3\% heavy hitter, and the model without 3\% light hitter.}
    \label{fig: abl_del_h2}
\vspace{-1em}
\end{wrapfigure}

After verifying the existence of heavy hitters,
we then mathematically and experimentally verify whether these heavy hitter neurons matter for the model performance or not.
For illustration convenience, we denote neurons with the lowest output norm as ``light hitters''.
As shown in Figure \ref{fig: abl_del_h2}, we uniformly remove the top-3\% ``heavy hitter'' and ``light hitter'' neurons at each layer from the model, respectively. Then we check the accuracy drop.
\textit{We observe that heavy hitters matter for model performance}. Specifically, removing top 3\% ``heavy hitter'' causes significant accuracy drop compared to removing ``light hitters''.

This phenomenon can also be understood mathematically: Suppose we remove $j^{\text{th}}$ neuron from the FFN.
$\mU'\in\mathbb{R}^{d \times (d_{ff}-1)}$ and $\mV'\in\mathbb{R}^{(d_{ff}-1) \times d}$ are obtained by removing the $j^{\text{th}}$ columns and rows from $\mU$ and $\mV$, respectively.
Given any input tensor $\mX$, the residual error of FFN output is:

\begin{align}
\|\sigma(\mX\mU)\mV-\sigma(\mX\mU')\mV'\|_F^2 
&= \|\sigma(\mX\mU_{:,j})\mV_{j,:}\|_F^2 \nonumber \\
&= \sum_{i}\sum_{k} \sigma(\mX\mU_{i,j})^2\mV_{j,k}^2 \nonumber \\
&= \sum_{i}\sigma(\mX\mU_{i,j})^2 (\sum_{k}\mV_{j,k}^2) \nonumber \\
&=\sum_{i}\sigma(\mX\mU_{i,j})^2 \|\mV_{j,:}\|\nonumber \nonumber\\
&=\|\sigma(\mX\mU_{:,j})\|_F^2 \|\mV_{j,:}\|_F^2. \label{eq: math_explain}
\end{align}

From \Eqref{eq: math_explain}, we can see that the residual error is controled by two terms, namely, the neuron output norm $\|\sigma(\mX\mU_{:,j})\|_F^2$ and $\|\mV_{j,:}\|_F^2$ which quantifies how much the neurons' outputs contribute to the FFN output.
According to Figure \ref{fig: h2 exists}, a few heavy hitter neurons have very large $\|\sigma(\mX\mU_{:,j})\|_F^2$.
Thus intuitively, if we remove them, the residual error must be much large than removing light hitter.

In the next section, we discuss how to utilize this observation for optimizing the trade-off between accuracy and efficiency.

\subsection{Framework}
\label{sec: implementation}

\begin{wrapfigure}{r}{0.36\textwidth}
 \vspace{-3em}
  \begin{center}
    \includegraphics[width=0.36\textwidth]{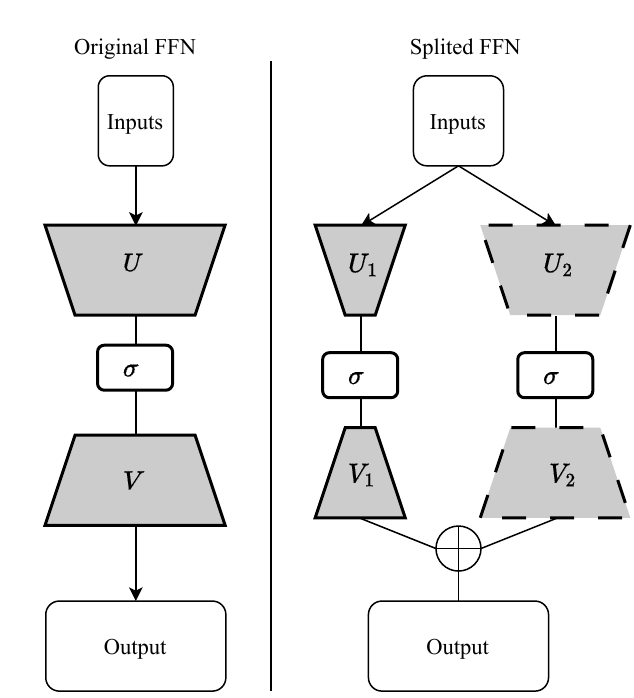}
  \end{center}
\vspace{-.5em}
    \caption{The diagram of our proposed method. We explicitly split the original FFN into two parts according to the set of heavy hitters $\texttt{h}_2$.
 $\mU_1=\mU_{:, \texttt{h}_2}$ and $\mV_1=\mV_{\texttt{h}_2, :}$.
Similarly, $\mU_2$ and $\mV_2$ are FFN weights specified by remain neuron.
We allow less resource to the FFN without heavy hitters, which is denoted with dotted lines.}
    \label{fig: diag}
\vspace{-5em}
\end{wrapfigure}

% As we shown in \Eqref{eq: ffn decompose}, the FFN module can be explicitly splitted into two parts according to the set of heavy hitters.
In Figure \ref{fig: diag} we present the overview of our framework.
The first step of our framework is to go over a small training set to identify which neuron is heavy hitter according to \Eqref{eq: math_explain}.
Then as we shown in \Eqref{eq: ffn decompose}, we explicitly split FFN module into two parts according to the set of heavy hitters.
We have experimentally shown that few heavy hitter neurons are crucial for the model accuracy.
Thus when applying compression methods,
our idea is to protect these few-but-important heavy hitters, namely, $\mU_1$ and $\mV_1$ in Figure \ref{fig: diag}.
For example, when applying low rank decomposition, we only decompose $\mU_2$ and $\mV_2$, while leaving $\mU_1$ and $\mV_1$ unchanged.

\section{Experiments}

In this Section, we combine the idea of \wta ~with different compression methods to improve their accuracy-efficiency trade-off on both Bert models and LLMs.

\subsection{Bert Experimental Analysis}
\subsubsection{Experimental Settings}

\paragraph{Datasets and Evaluation Protocol.} 
Following most of the previous work, we adopt GLUE benchmark~\citep{wang2018glue} to evaluate the effectiveness of different methods, including the CoLA, SST-2, MRPC, QQP, MNLI, QNLI, and RTE datasets.
For the SST-2, MNLI, QNLI, and RTE datasets, we report the validation accuracy. For CoLA, we use Matthew's correlation as the evaluation metric. The F1 score is reported for both MRPC and QQP tasks.
All reported numbers are averaged over three random trials.

\paragraph{Adopted Models and Compression Methods.} 
For the backbone model, we follow the previous work to adopt the Bert-Base \citep{devlin2018bert} and Bert-Large for evaluating the effectiveness of different methods. Here we only apply low-rank decomposition to $\mU_2$ and $\mV_2$ in Figure \ref{fig: diag}, while leaving $\mU_1$ and $\mV_1$ unchanged. 

\paragraph{Hyperparameter Settings.}

For Bert model, we preserve top $25\%$ heavy hitter neurons measured by the importance score defined in \Eqref{eq: math_explain}. 
For the remain part, we apply low rank decomposition with Singular Value Decomposition (SVD). Specifically, we use a rank that is $10\%$ of the full rank.
For a fair comparison, we compare \wta ~against the vanilla SVD under the same parameter budget.
We note that we further fine-tune the compressed Bert for a few epochs.
We note that in this setting, our method can reduce total model parameters by 43.1\% (excluding embedding).

\subsubsection{Accuracy-Efficiency Trade-Off}

We first test our idea on GLUE dataset with Bert-base and Bert-large.
As shown in Table \ref{tab: exp res}, we observe that 
\ding{182} Low rank decomposition with \wta~ significantly outperforms vanilla low rank decomposition under the same parameter budget.
Specifically, when our method was applied to Bert-base, there was an accuracy decrease of 0.3\%, and a 1.0\% drop for Bert-large. In comparison, the standard low-rank decomposition resulted in a 1.6\% accuracy drop for Bert-base and a significant 5.1\% drop for Bert-large.
As we analyzed, a few heavy hitter neurons are significantly important than the other neurons in terms of the impact to performance.
Thus, applying vanilla low rank decomposition to all neurons will destroy this structure.
In Table \ref{tab: speed}, we report the wall clock inference time of Bert-base with our method on commodity hardware such as CPUs and GPUs. 
We observe \ding{183}  Low rank decomposition with \wta ~is $1.25\sim1.56\times$ faster than the baseline, depending on the inference word load.
Here in Table \ref{tab: speed} we do not include the vanilla low rank decomposition because its accuracy drop is not acceptable, let alone the efficiency.

\begin{table}
\centering
\caption{The experimental comparison between \wta ~and vanilla low rank decomposition. All reported results are averaged over three random trials.}
\label{tab: exp res}
\resizebox{\textwidth}{!}{
\begin{tblr}{
  cells = {c},
  cell{2}{1} = {r=3}{},
  cell{5}{1} = {r=3}{},
  hline{1-2} = {-}{},
  hline{8} = {1-10}{},
}
Model      & Method     & Cola     & RTE      & MRPC (F1) & SST2     & QNLI     & MNLI     & QQP      & Average \\
Bert-base  & Baseline   & 57±1.0       & 63.2±0.2     & 89.2±0.5      & 93±0.3       & 91.6±0.1     & 85±0.1       & 90.8±0.0     & 81.4    \\
           & \wta (Low Rank) & 56.3±0.3 & 65.8±0.2 & 89.7±0.5  & 91.8±0.4 & 90.3±0.1 & 83.2±0.2 & 90.8±0.1 & 81.1    \\
           & Low Rank   & 44.3±1.0 & 62.3±0.2 & 86.0±0.8  & 91.2±0.3 & 89.0±0.1 & 82.4±0.1 & 90.8±0.0 & 79.8    \\
Bert-large & Baseline   & 60.3±0.3     & 69.7±0.4     & 90.6±0.2      & 93.7±0.3     & 92.4±0.2     & 86.6±0.2     & 91.4±0.0     & 83.5    \\
           & \wta (Low Rank) & 56.7±0.7 & 71.8±0.3 & 89.6±0.1  & 92.2±0.1 & 91.4±0.1 & 84.8±0.0 & 91.2±0.0 & 82.5    \\
           & Low Rank   &   3.7±5.2       &     53.2±0.5     &  84.4±2.3         &    91.2±0.3      &   88.1±1.2       &   84.4±0.3       &    91.1±0.2      &         78.4
\end{tblr}
}
\end{table}

\begin{table}
\centering
\caption{Inference speed (ms) on both CPU and GPU. Here ``BS'' refers to the batch size and ``Seq. Length'' is the sequential length of the input texts. \wta~(Low Rank) can have $1.25\sim 1.56\times$ wall clock time speedup on commodity hardware.}
\label{tab: speed}
\resizebox{\textwidth}{!}{
\begin{tabular}{ccccc|cccc} 
\hline
Hardware                       & \multicolumn{4}{c|}{NVIDIA V100}                                           & \multicolumn{4}{c}{Intel CPU E5-2699A}                                     \\ 
\hline
\multirow{2}{*}{Configuration} & \multicolumn{2}{c}{Seq. Length=128} & \multicolumn{2}{c|}{Seq. Length=256} & \multicolumn{2}{c}{Seq. Length=128} & \multicolumn{2}{c}{Seq. Length=256}  \\
                               & BS=8   & BS=32                      & BS=8  & BS=32                        & BS=8   & BS=32                      & BS=8   & BS=32                       \\ 
\hline
Baseline                       & 19.1  & 68.2                     & 36.7 & 137.1                       & 275.4 & 1211.4                    & 647.2 & 3214.2                     \\
\wta~(Low Rank)                            &  15.2 (1.25$\times$) & 51.5 (1.32$\times$)                      & 29.7 (1.24$\times$)   & 104.6 (1.24$\times$)                    & 207 (1.33$\times$)  & \multicolumn{1}{c}{777.4} (1.56$\times$) & 514.1 (1.26$\times$) & 2571 (1.25$\times$)               \\
\hline
\end{tabular}}
\end{table}

\subsection{LLM Results}

\begin{table}[h!]
\centering
\caption{The experimental comparison between \wta~ and vanilla round-to-nearest (RTN) and AWQ quantization. ``w3-g128'' refers 3-bit weight quantization with a group size 128.}
\label{tab: opt-res}
\begin{tabular}{cccc} 
\hline
\multicolumn{1}{l}{Wikitext2 PPL $\downarrow$} & \multicolumn{1}{l}{} & \multicolumn{1}{l}{OPT-1.3B} & \multicolumn{1}{l}{OPT-6.7B}  \\ 
\hline
FP16                                           & -                    & 14.62                        & 12.29                         \\ 
\hline
\multirow{4}{*}{INT3-g128}                     & RTN                  & 207.4                        & 43.16                         \\
                                               & RTN+\wta                 & 81.4                         & 23.88                         \\
                                               & AWQ                  & 18.53                        & 12.99                         \\
                                               & AWQ+\wta                & 18.33                        & 12.90                         \\
\hline
\end{tabular}
\end{table}

% \subsubsection{Experimental Settings}
Here we integrate \wta~ with the vanilla round-to-nearest quantization to compress the OPT model \cite{opt}.
We choose round-to-nearest quantization mainly because it is a strong baseline when using a small group size like 128 \cite{lin2023awq}. 
Here we examine our idea on both OPT-1.3B and OPT-6.7B.
We use 8-bit quantization for all heavy hitter neurons with a group size 128, while all other parts are quantized into 3-bit with a group size 128.
The results are shown in Table \ref{tab: opt-res}.
We observe that \wta~ significantly outperforms the vanilla quantization.

\section{Conclusion}

Optimizing the efficiency-accuracy is essential for effectively deploying LLMs on commodity hardware.
A significant portion of the efficiency challenge is the Feed-
forward network (FFN) component, which accounts for roughly $\frac{2}{3}$ total parameters and inference latency.
In this paper, we first observe that only a few neurons of FFN module have large output norm for any
input tokens, while the others are sparsely triggered by different tokens. Based on this observation, we
explicitly split the FFN into two parts according to the heavy hitters. We improve the efficiency-accuracy
trade-off of existing compression methods by allocating more resource to FFN parts with heavy hitters.

\bibliographystyle{plain}
\bibliography{ref}

\begin{thebibliography}{10}

\bibitem{gpt3}
Tom Brown, Benjamin Mann, Nick Ryder, Melanie Subbiah, Jared~D Kaplan, Prafulla
  Dhariwal, Arvind Neelakantan, Pranav Shyam, Girish Sastry, Amanda Askell,
  et~al.
\newblock Language models are few-shot learners.
\newblock {\em Advances in neural information processing systems},
  33:1877--1901, 2020.

\bibitem{brown2020language}
Tom Brown, Benjamin Mann, Nick Ryder, Melanie Subbiah, Jared~D Kaplan, Prafulla
  Dhariwal, Arvind Neelakantan, Pranav Shyam, Girish Sastry, Amanda Askell,
  et~al.
\newblock Language models are few-shot learners.
\newblock {\em Advances in neural information processing systems},
  33:1877--1901, 2020.

\bibitem{chen2021drone}
Patrick Chen, Hsiang-Fu Yu, Inderjit Dhillon, and Cho-Jui Hsieh.
\newblock Drone: Data-aware low-rank compression for large nlp models.
\newblock {\em Advances in neural information processing systems},
  34:29321--29334, 2021.

\bibitem{dettmers2022llm}
Tim Dettmers, Mike Lewis, Younes Belkada, and Luke Zettlemoyer.
\newblock Llm. int8 (): 8-bit matrix multiplication for transformers at scale.
\newblock {\em arXiv preprint arXiv:2208.07339}, 2022.

\bibitem{devlin2018bert}
Jacob Devlin, Ming-Wei Chang, Kenton Lee, and Kristina Toutanova.
\newblock Bert: Pre-training of deep bidirectional transformers for language
  understanding.
\newblock {\em arXiv preprint arXiv:1810.04805}, 2018.

\bibitem{frantar2023sparsegpt}
Elias Frantar and Dan Alistarh.
\newblock Sparsegpt: Massive language models can be accurately pruned in
  one-shot.
\newblock 2023.

\bibitem{frantar2022gptq}
Elias Frantar, Saleh Ashkboos, Torsten Hoefler, and Dan Alistarh.
\newblock Gptq: Accurate post-training quantization for generative pre-trained
  transformers.
\newblock {\em arXiv preprint arXiv:2210.17323}, 2022.

\bibitem{mlcllm}
GitHub.
\newblock \url{https://github.com/mlc-ai/mlc-llm}, 2023.

\bibitem{web-llm}
GitHub.
\newblock \url{https://github.com/mlc-ai/web-llm}, 2023.

\bibitem{he2018amc}
Yihui He, Ji~Lin, Zhijian Liu, Hanrui Wang, Li-Jia Li, and Song Han.
\newblock Amc: Automl for model compression and acceleration on mobile devices.
\newblock In {\em Proceedings of the European conference on computer vision
  (ECCV)}, pages 784--800, 2018.

\bibitem{hendrycks2016gaussian}
Dan Hendrycks and Kevin Gimpel.
\newblock Gaussian error linear units (gelus).
\newblock {\em arXiv preprint arXiv:1606.08415}, 2016.

\bibitem{hubara2021accelerated}
Itay Hubara, Brian Chmiel, Moshe Island, Ron Banner, Joseph Naor, and Daniel
  Soudry.
\newblock Accelerated sparse neural training: A provable and efficient method
  to find n: m transposable masks.
\newblock {\em Advances in Neural Information Processing Systems},
  34:21099--21111, 2021.

\bibitem{kaplan2020scaling}
Jared Kaplan, Sam McCandlish, Tom Henighan, Tom~B Brown, Benjamin Chess, Rewon
  Child, Scott Gray, Alec Radford, Jeffrey Wu, and Dario Amodei.
\newblock Scaling laws for neural language models.
\newblock {\em arXiv preprint arXiv:2001.08361}, 2020.

\bibitem{kwon2022fast}
Woosuk Kwon, Sehoon Kim, Michael~W Mahoney, Joseph Hassoun, Kurt Keutzer, and
  Amir Gholami.
\newblock A fast post-training pruning framework for transformers.
\newblock {\em arXiv preprint arXiv:2204.09656}, 2022.

\bibitem{vllm}
Woosuk Kwon, Zhuohan Li, Siyuan Zhuang, Ying Sheng, Lianmin Zheng, Cody~Hao Yu,
  Joseph~E Gonzalez, Hao Zhang, and Ion Stoica.
\newblock Efficient memory management for large language model serving with
  pagedattention.
\newblock {\em arXiv preprint arXiv:2309.06180}, 2023.

\bibitem{lazyneuron}
Zonglin Li, Chong You, Srinadh Bhojanapalli, Daliang Li, Ankit~Singh Rawat,
  Sashank~J. Reddi, Ke~Ye, Felix Chern, Felix Yu, Ruiqi Guo, and Sanjiv Kumar.
\newblock The lazy neuron phenomenon: On emergence of activation sparsity in
  transformers.
\newblock In {\em The Eleventh International Conference on Learning
  Representations}, 2023.

\bibitem{lin2023awq}
Ji~Lin, Jiaming Tang, Haotian Tang, Shang Yang, Xingyu Dang, and Song Han.
\newblock Awq: Activation-aware weight quantization for llm compression and
  acceleration.
\newblock {\em arXiv preprint arXiv:2306.00978}, 2023.

\bibitem{dejavu}
Zichang Liu, Jue Wang, Tri Dao, Tianyi Zhou, Binhang Yuan, Zhao Song, Anshumali
  Shrivastava, Ce~Zhang, Yuandong Tian, Christopher Re, et~al.
\newblock Deja vu: Contextual sparsity for efficient llms at inference time.
\newblock In {\em International Conference on Machine Learning}, pages
  22137--22176. PMLR, 2023.

\bibitem{liu2023DejaVu}
Zichang Liu, Jue Wang, Tri Dao, Tianyi Zhou, Binhang Yuan, Zhao Song, Anshumali
  Shrivastava, Ce~Zhang, Yuandong Tian, Christopher Ré, and Beidi Chen.
\newblock Deja vu: Contextual sparsity for efficient llms at inference time.
\newblock In {\em International Conference on Machine Learning}. PMLR, 2023.

\bibitem{wtacrs}
Zirui Liu, Guanchu Wang, Shaochen Zhong, Zhaozhuo Xu, Daochen Zha, Ruixiang
  Tang, Zhimeng Jiang, Kaixiong Zhou, Vipin Chaudhary, Shuai Xu, et~al.
\newblock Winner-take-all column row sampling for memory efficient adaptation
  of language model.
\newblock {\em arXiv preprint arXiv:2305.15265}, 2023.

\bibitem{liu2023winner}
Zirui Liu, Guanchu Wang, Shaochen Zhong, Zhaozhuo Xu, Daochen Zha, Ruixiang
  Tang, Zhimeng Jiang, Kaixiong Zhou, Vipin Chaudhary, Shuai Xu, et~al.
\newblock Winner-take-all column row sampling for memory efficient adaptation
  of language model.
\newblock {\em arXiv preprint arXiv:2305.15265}, 2023.

\bibitem{nagel2020up}
Markus Nagel, Rana~Ali Amjad, Mart Van~Baalen, Christos Louizos, and Tijmen
  Blankevoort.
\newblock Up or down? adaptive rounding for post-training quantization.
\newblock In {\em International Conference on Machine Learning}, pages
  7197--7206. PMLR, 2020.

\bibitem{radford2018improving}
Alec Radford, Karthik Narasimhan, Tim Salimans, Ilya Sutskever, et~al.
\newblock Improving language understanding by generative pre-training.
\newblock 2018.

\bibitem{radford2019language}
Alec Radford, Jeffrey Wu, Rewon Child, David Luan, Dario Amodei, Ilya
  Sutskever, et~al.
\newblock Language models are unsupervised multitask learners.
\newblock {\em OpenAI blog}, 1(8):9, 2019.

\bibitem{t5}
Colin Raffel, Noam Shazeer, Adam Roberts, Katherine Lee, Sharan Narang, Michael
  Matena, Yanqi Zhou, Wei Li, and Peter~J Liu.
\newblock Exploring the limits of transfer learning with a unified text-to-text
  transformer.
\newblock {\em The Journal of Machine Learning Research}, 21(1):5485--5551,
  2020.

\bibitem{swiglu}
Noam Shazeer.
\newblock Glu variants improve transformer.
\newblock {\em arXiv preprint arXiv:2002.05202}, 2020.

\bibitem{sheng2023high}
Ying Sheng, Lianmin Zheng, Binhang Yuan, Zhuohan Li, Max Ryabinin, Daniel~Y Fu,
  Zhiqiang Xie, Beidi Chen, Clark Barrett, Joseph~E Gonzalez, and othersi.
\newblock High-throughput generative inference of large language models with a
  single gpu.
\newblock In {\em International Conference on Machine Learning}. PMLR, 2023.

\bibitem{transformer}
Ashish Vaswani, Noam Shazeer, Niki Parmar, Jakob Uszkoreit, Llion Jones,
  Aidan~N Gomez, {\L}ukasz Kaiser, and Illia Polosukhin.
\newblock Attention is all you need.
\newblock {\em Advances in neural information processing systems}, 30, 2017.

\bibitem{wang2018glue}
Alex Wang, Amanpreet Singh, Julian Michael, Felix Hill, Omer Levy, and Samuel~R
  Bowman.
\newblock Glue: A multi-task benchmark and analysis platform for natural
  language understanding.
\newblock {\em arXiv preprint arXiv:1804.07461}, 2018.

\bibitem{cofi}
Mengzhou Xia, Zexuan Zhong, and Danqi Chen.
\newblock Structured pruning learns compact and accurate models.
\newblock {\em arXiv preprint arXiv:2204.00408}, 2022.

\bibitem{xiao2022smoothquant}
Guangxuan Xiao, Ji~Lin, Mickael Seznec, Julien Demouth, and Song Han.
\newblock Smoothquant: Accurate and efficient post-training quantization for
  large language models.
\newblock {\em arXiv preprint arXiv:2211.10438}, 2022.

\bibitem{xiao2023smoothquant}
Guangxuan Xiao, Ji~Lin, Mickael Seznec, Hao Wu, Julien Demouth, and Song Han.
\newblock Smoothquant: Accurate and efficient post-training quantization for
  large language models.
\newblock In {\em International Conference on Machine Learning}, pages
  38087--38099. PMLR, 2023.

\bibitem{xu2023compress}
Zhaozhuo Xu, Zirui Liu, Beidi Chen, Yuxin Tang, Jue Wang, Kaixiong Zhou, Xia
  Hu, and Anshumali Shrivastava.
\newblock Compress, then prompt: Improving accuracy-efficiency trade-off of llm
  inference with transferable prompt.
\newblock {\em arXiv preprint arXiv:2305.11186}, 2023.

\bibitem{opt}
Susan Zhang, Stephen Roller, Naman Goyal, Mikel Artetxe, Moya Chen, Shuohui
  Chen, Christopher Dewan, Mona Diab, Xian Li, Xi~Victoria Lin, et~al.
\newblock Opt: Open pre-trained transformer language models.
\newblock {\em arXiv preprint arXiv:2205.01068}, 2022.

\bibitem{zhao2023inrank}
Jiawei Zhao, Yifei Zhang, Beidi Chen, Florian Sch{\"a}fer, and Anima
  Anandkumar.
\newblock Inrank: Incremental low-rank learning.
\newblock {\em arXiv preprint arXiv:2306.11250}, 2023.

\end{thebibliography}

% \section{Appendix}

\end{document}